%% file: paper.tex
\documentclass[runningheads]{llncs}
\usepackage[T1]{fontenc}
\usepackage{graphicx}
\usepackage{booktabs}
\usepackage[misc]{ifsym}
\newcommand{\corr}{(\Letter)}
\usepackage{mwe}
\usepackage{amsmath}
\usepackage{comment}
\usepackage{todonotes}
\usepackage{hyperref}
\usepackage{pifont}
\usepackage{subcaption}
\begin{document}

\title{Adversarial Robustness of Variational Autoencoders across Intersectional Subgroups
}

\titlerunning{Adversarial Robustness of VAEs across Intersectional Subgroups}

\author{Author information scrubbed for double-blind reviewing}
 \author{Chethan Krishnamurthy Ramanaik\corr\orcidID{0009-0002-6718-2665} \and
 Arjun Roy\orcidID{0000-0002-4279-9442}  \and
 Eirini Ntoutsi\orcidID{0000-0001-5729-1003}}

\authorrunning{C.K. Ramanaik et al.}

 \institute{University of the Bundeswehr Munich, Germany \email{\{chethan.krishnamurthy,arjun.roy,eirini.ntoutsi\}@unibw.de}
}

\maketitle              

\begin{abstract}

Despite advancements in Autoencoders (AEs) for tasks like dimensionality reduction, representation learning and data generation, they remain vulnerable to adversarial attacks. Variational Autoencoders (VAEs), with their probabilistic approach to disentangling latent spaces, show stronger resistance to such perturbations compared to deterministic AEs; however, their resilience against adversarial inputs is still a concern.
This study evaluates the robustness of VAEs against non-targeted adversarial attacks by optimizing minimal sample-specific perturbations to cause maximal damage across diverse demographic subgroups (combinations of age and gender).
We investigate two questions: whether there are robustness disparities among subgroups, and what factors contribute to these disparities, such as data scarcity and representation entanglement. 
Our findings reveal that robustness disparities exist but are not always correlated with the size of the subgroup.
By using downstream gender and age classifiers and examining latent embeddings, we highlight the vulnerability of subgroups like older women, who are prone to misclassification due to adversarial perturbations pushing their representations toward those of other subgroups.


\end{abstract}

\section{Introduction}
\label{sec:intro}
\input{intro}

\section{Related Work}
\label{sec:related}
\input{related}

\section{Basic concepts and problem formulation}
\label{sec.basics}
\input{background}
\section{Evaluating adversarial robustness across subgroups}
\label{sec:our}
\input{method}

\section{Experiments}
\label{sec.exp}
We evaluate the robustness of different $\beta$-VAE models on different demographic subgroups of the CelebA dataset (Section~\ref{sec:expSetup}).  We report both quantitative (Section~\ref{sec:expQuantitative}) and qualitative (Section~\ref{sec:expQualitative}) results.

\subsection{Experimental setup}
\label{sec:expSetup}
\subsubsection{Dataset}
\label{sec:dataset}
\input{dataset}
\subsubsection{Training}
Adam optimizer is used for training with a learning rate of 1e-4.
We train three different VAE models, with $\beta = \{1, 5, 10\}$. Higher value of $\beta$ refers to higher emphasis of the VAE to the disentanglement factor (c.f. Section~\ref{sec:betaVAE}) in the latent space. However, when $\beta$ is too low or too high the model learns an entangled latent representation due to either too much or too little capacity in the latent z bottleneck ~\cite{higgins2017beta}.
The full code for all the experiments and evaluation along with additional resources can be accessed at \href{https://github.com/ChethanKodase/robustness_of_subgroups}{\url{https://github.com/ChethanKodase/robustness_of_subgroups}}

\subsubsection{Evaluation setting and evaluation measures}
To evaluate the robustness of the subgroups, we selected 60 random samples from each subgroup. This decision was driven by resource constraints, as generating sample-specific white-box adversarial attacks incurs significant computational overhead.
We evaluate the robustness of VAEs to samples of different groups by measuring the adversarial deviations according to Equation~\ref{adv_div}. For all the trained models and chosen random samples we optimized adversarial perturbations according to Equation~\ref{max_damage} by empirically setting the perturbation norm bound \emph{c} for all experiments to enhance robustness comparisons. A very low \emph{c} may cause no output damage in some samples, while an extreme \emph{c} could cause excessive damage in all output samples, both of which hinder effective comparison. Our work aims to evaluate and compare adversarial robustness, not to demonstrate attack success. Since autoencoders are more resilient than classifiers \cite{gondim2018adversarial}, we use perceptibly intense perturbations, as also seen in \cite{willetts2019improving,gondim2018adversarial,kuzina2022alleviating}, to enable robustness comparisons. Then, we report on the distribution of adversarial deviation $\Delta_c$ for different subgroups, according to Equation~\ref{adv_div}.  
\begin{figure}
    \centering    \includegraphics[width=0.5\textwidth]{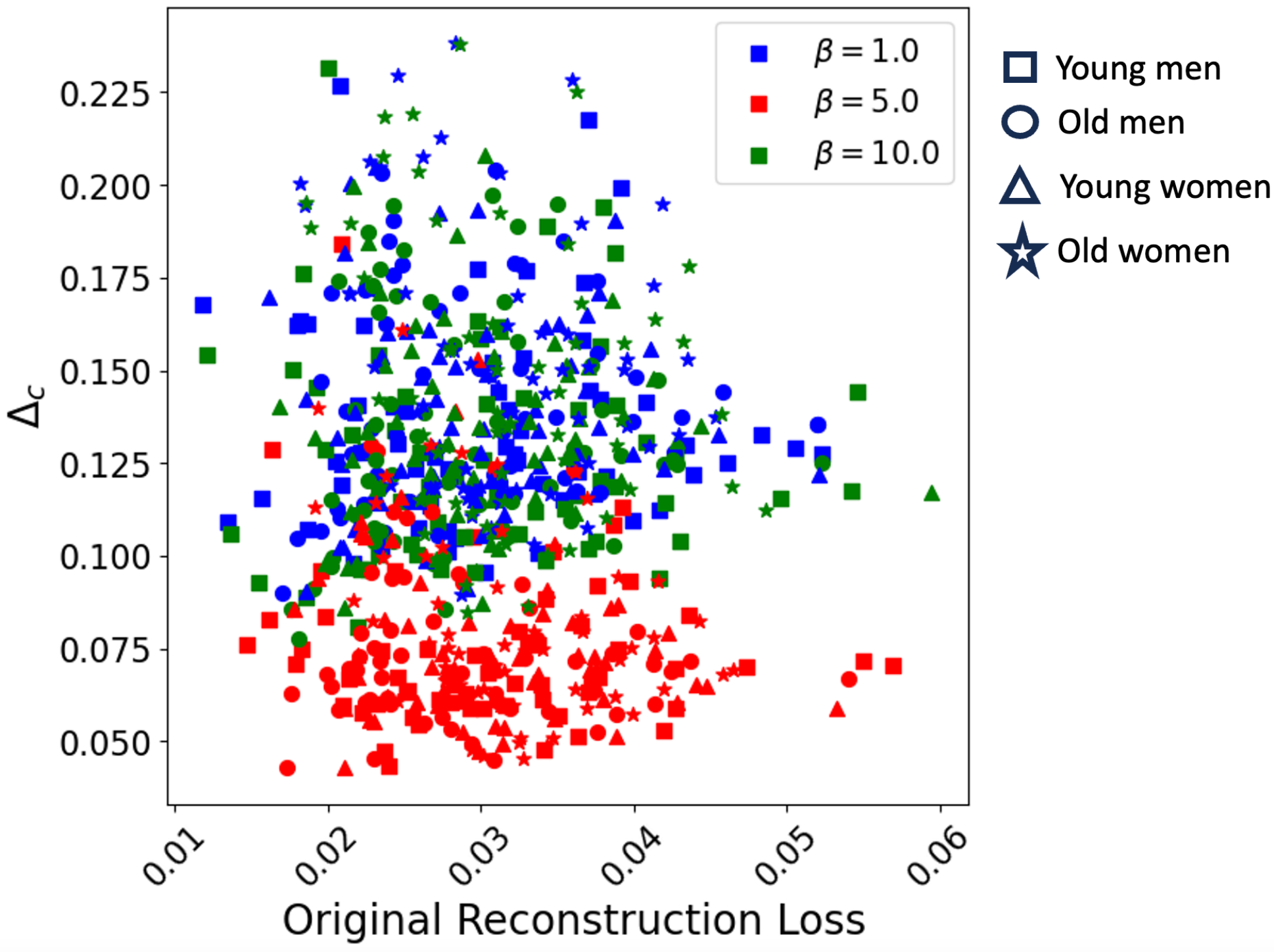}
    \caption{\small Adversarial deviation vs original (unperturbed) reconstruction loss for different subgroup instances (subgroup denoted by symbol) and different $\beta$-VAEs (denoted by color).  The lower the adversarial deviation the higher the robustness.  
    }
    \label{fig:beta_vae_scatter}
\end{figure}

In Figure ~\ref{fig:beta_vae_scatter} we show the adversarial deviation (c.f., Equation~\ref{adv_div}) vs unperturbed reconstruction loss (eq.~\ref{eq.loss_vae}) of VAE's with different $\beta$=1,5,and 10, across different subgroup instances. 
Higher values of adversarial deviation correspond to lower adversarial robustness of the model against the chosen samples for robustness evaluation against a given model. Lower values of unperturbed reconstruction loss correspond to better reconstruction of the unperturbed image. We notice that the adversarial deviations of samples are notably higher for attacks against VAE with $\beta=1$ and $\beta=10$ compared to that with $\beta=5$. This observation falls inline with the literature suggesting neither low or high values of $\beta$ are good~\cite{higgins2017beta}. Interestingly, we also notice that samples with relatively higher reconstruction loss (0.05 - 0.06) in the unperturbed scenario (mostly populated with young men and women) does not have higher deviance ($\Delta_c\leq 0.125$) compared to some samples (old men and women) with lower unperturbed reconstruction loss (0.02 - 0.03) with $\Delta_c\geq 0.2$. Reconstruction loss is generally evaluated as the proxy of models quality in understanding the latent feature semantics. This finding reveals the vulnerability of certain subgroups where, despite the low reconstruction loss of the original image, the model fails to comprehend the feature space and resist adversarial perturbations against samples belonging to these subgroups.

\section{Evaluation results}
\subsection{Quantitative analysis}
\label{sec:expQuantitative}
\begin{figure}
    \centering  
    \begin{subfigure}[b]{0.49\textwidth}
        \includegraphics[width=0.49\textwidth]{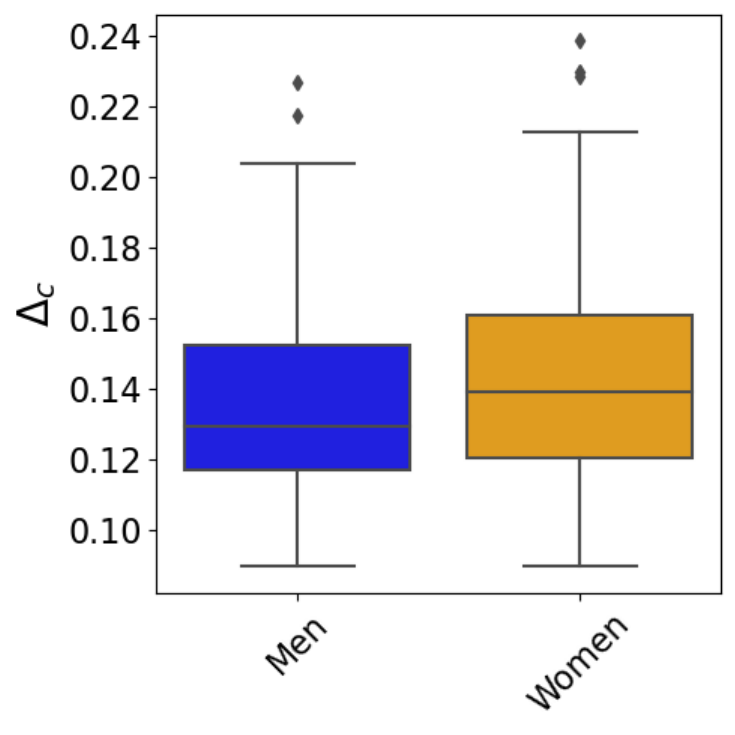}
        \includegraphics[width=0.49\textwidth]{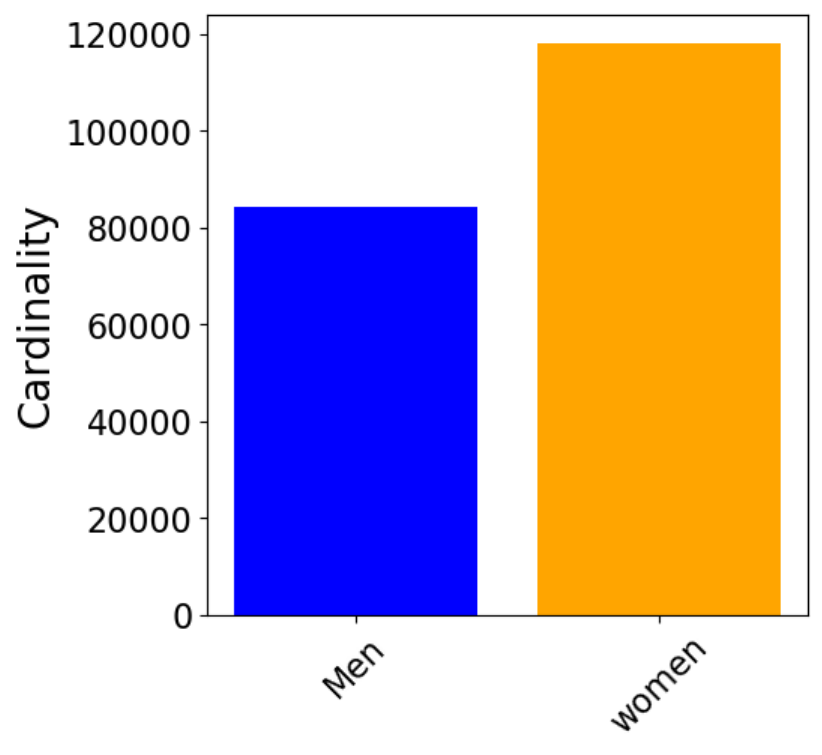}
        \caption{Gender}
    \end{subfigure}
\begin{subfigure}[b]{0.49\textwidth}
        \includegraphics[width=0.49\textwidth]{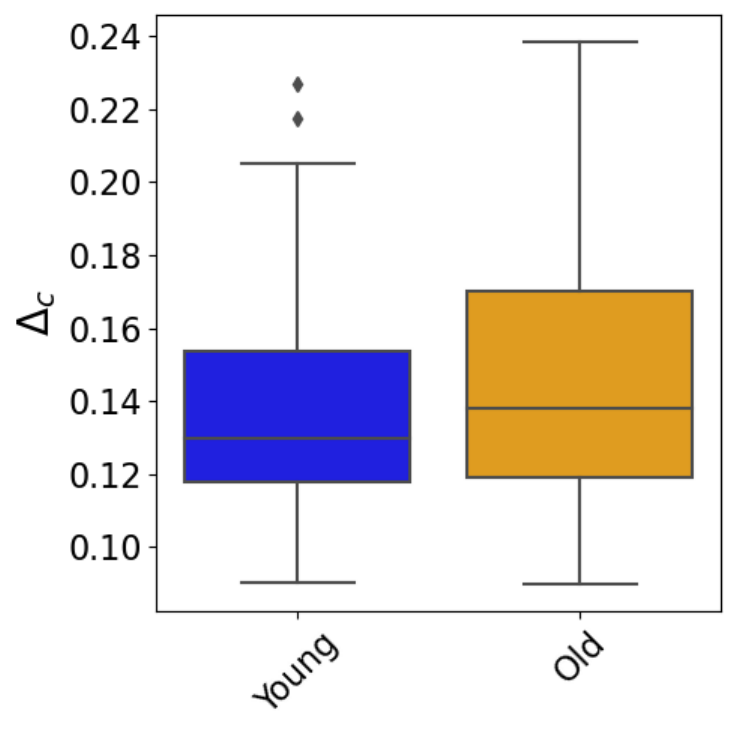}   
        \includegraphics[width=0.49\textwidth]{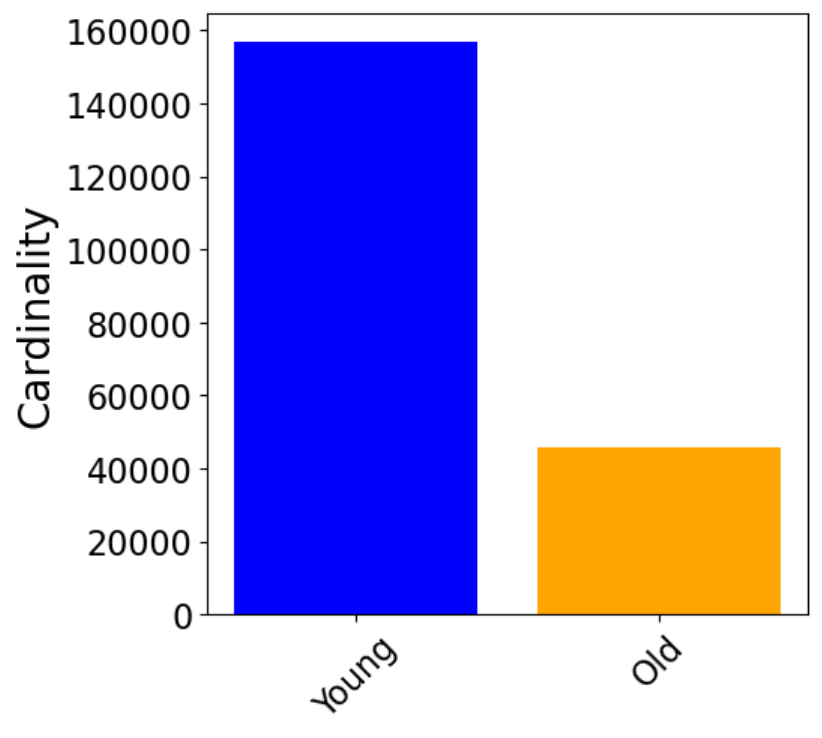}
        \caption{Age}
    \end{subfigure}
    \caption{\small Distribution of adversarial deviations for Gender and Age along with the group cardinalities.}
    \label{fig:gender_age_report}
\end{figure}
\begin{figure}
\vspace{-1mm}
    \centering
    \begin{subfigure}[b]{0.33\textwidth}
        \includegraphics[width=\textwidth]{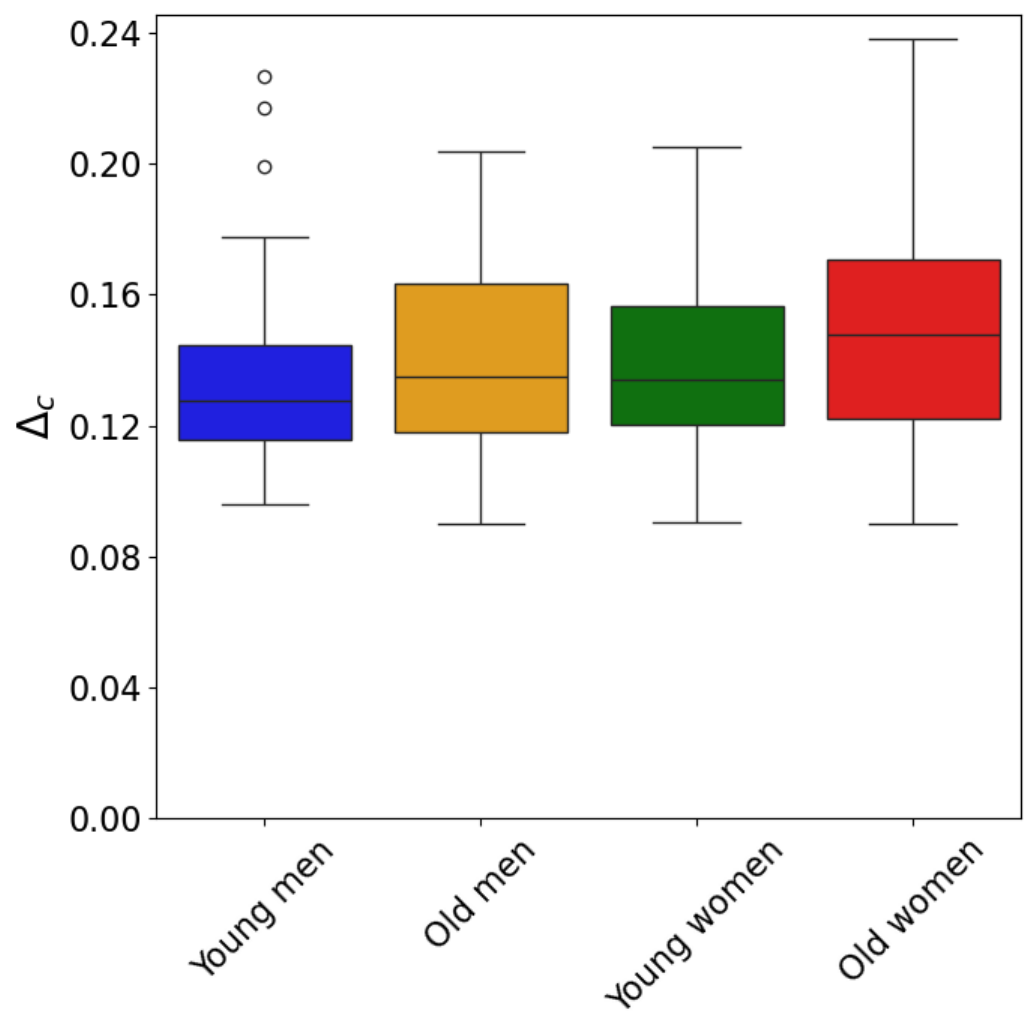}
        \caption{$\beta=1$}\label{fig.vanillaSubG}
    \end{subfigure}
    \begin{subfigure}[b]{0.33\textwidth}
        \includegraphics[width=\textwidth]{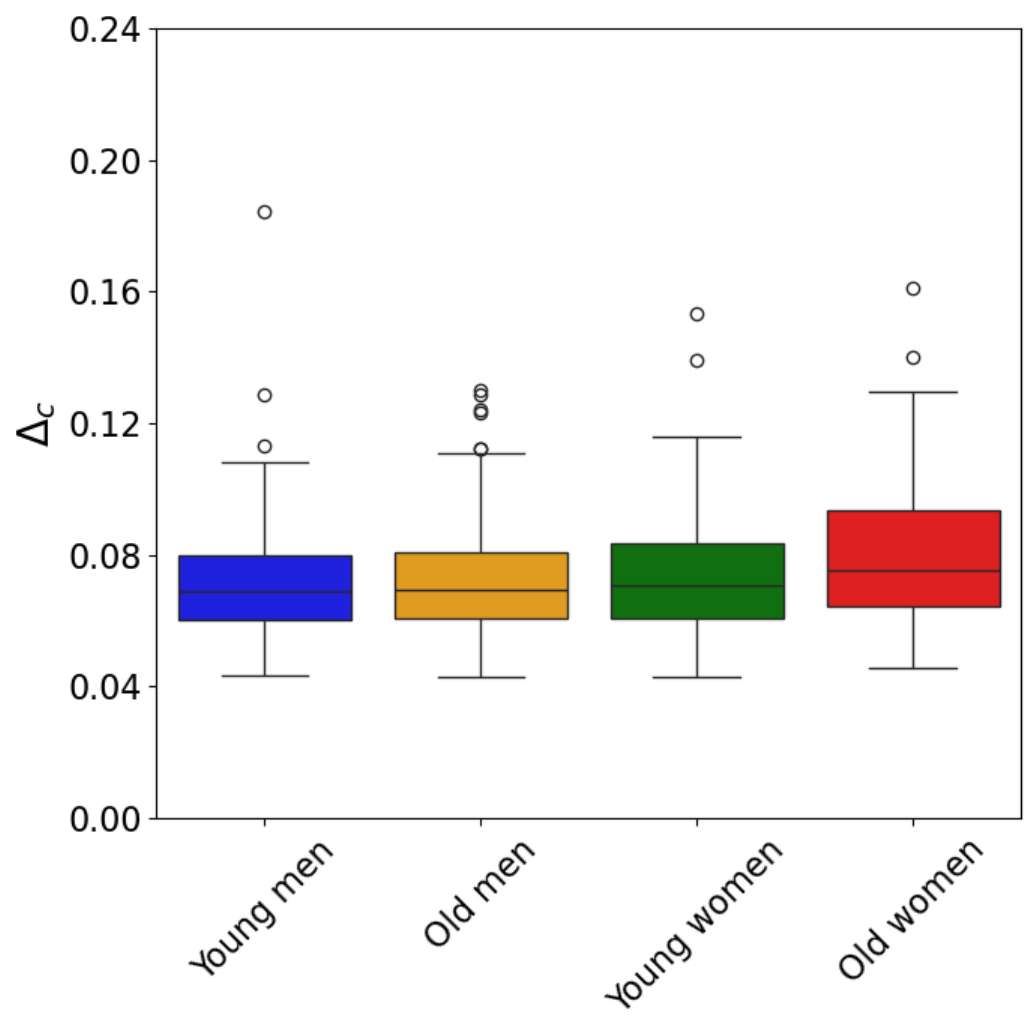}
        \caption{$\beta=5$}\label{fig.beta5}
    \end{subfigure}
    \hfill\\
    \begin{subfigure}[b]{0.33\textwidth}
         \includegraphics[width=\textwidth]{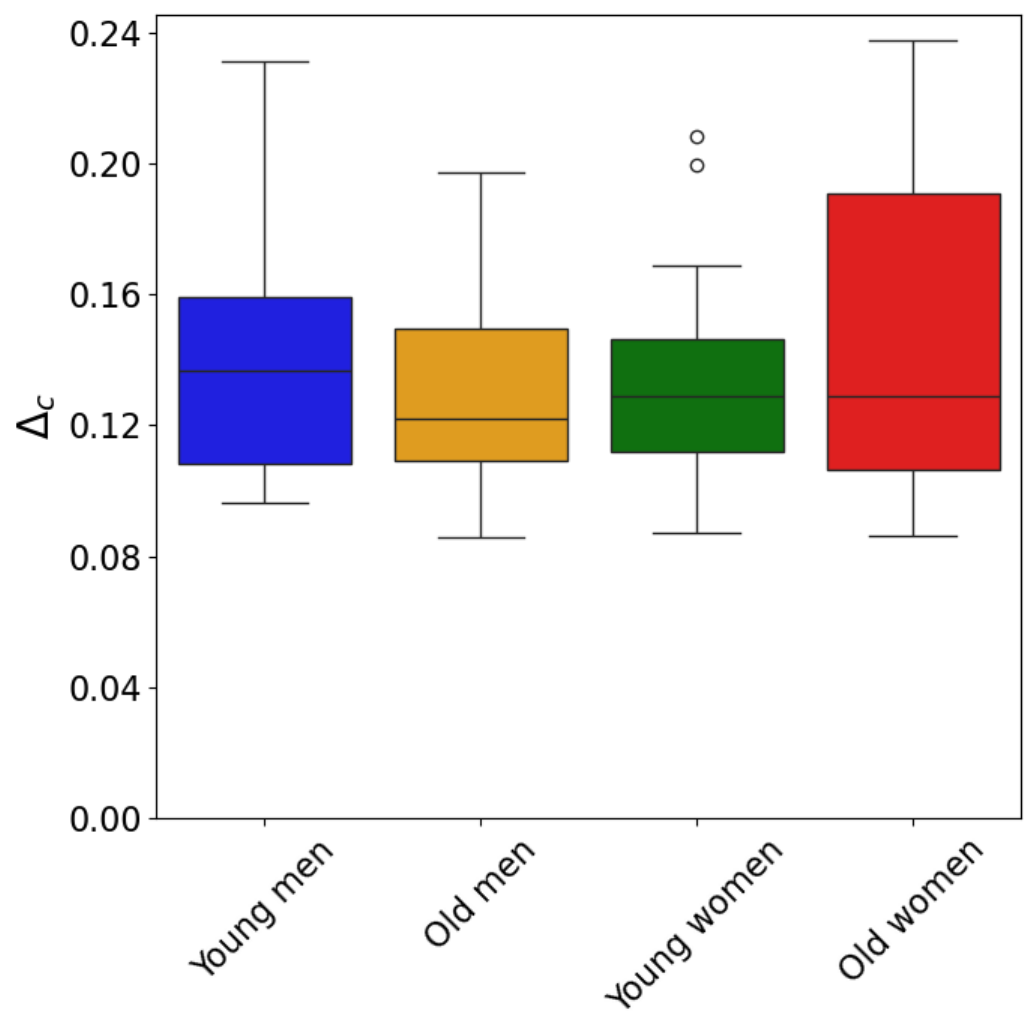}
         \caption{$\beta=10$}\label{fig.beta10}
    \end{subfigure}
    \begin{subfigure}[b]{0.33\textwidth}
         \includegraphics[width=\textwidth]{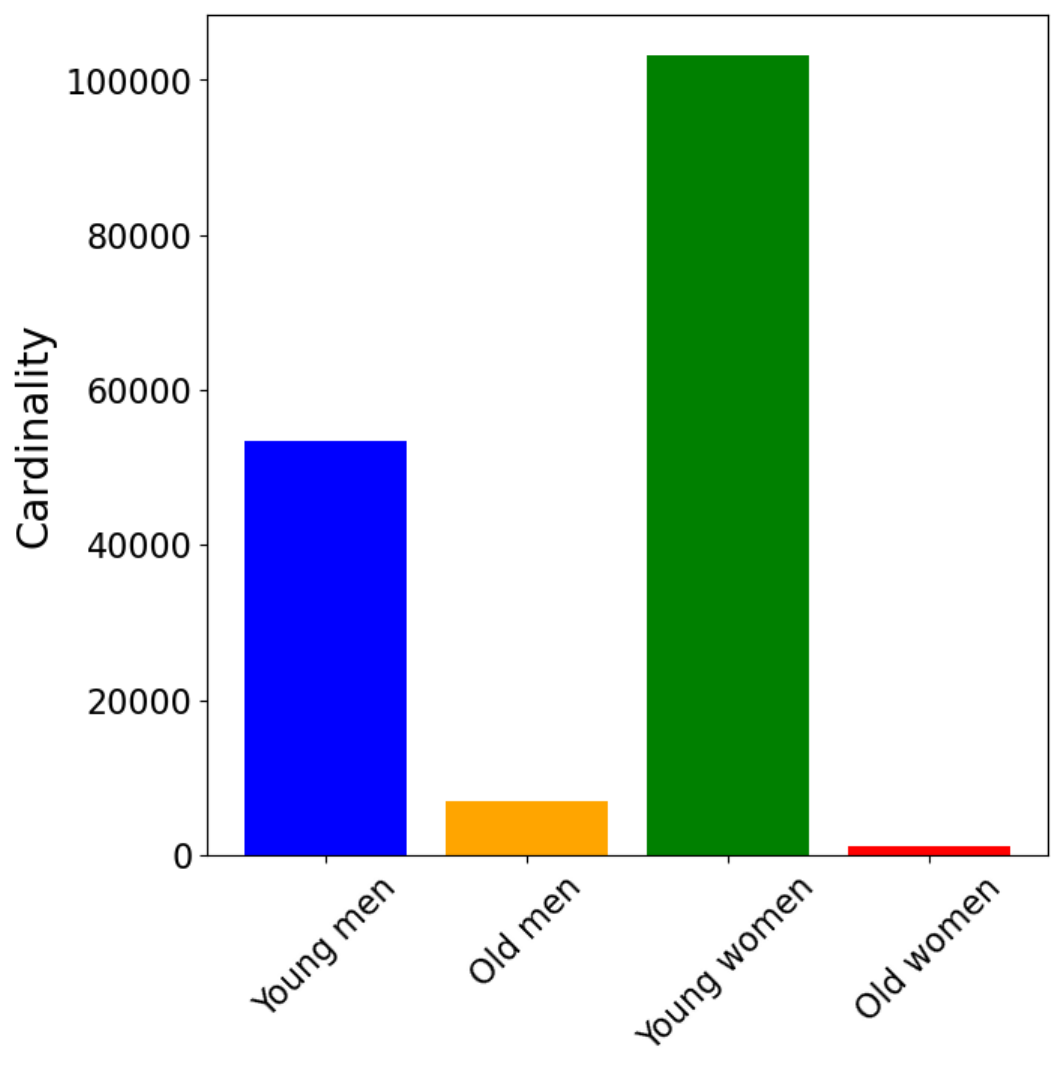}
         \caption{Subgroup cardinality}\label{fig.subGcard}
    \end{subfigure}    
    \caption{\small Distributions of adversarial robustness of vanilla VAE ($\beta=1$), and  $\beta$-VAE on subgroups defined by age and gender.}
\label{fig:instability_boxplot1}
\vspace{-8mm}
\end{figure}
\subsubsection{Adversarial robustness across gender and age groups}
We first report on the  adversarial deviations of gender and age groups alongside their respective cardinalities, in Figure~\ref{fig:gender_age_report}. 
Looking at the Age attribute, we can see that the \emph{Old} group exhibits relatively higher adversarial deviation (higher median and higher variance), indicative of lower adversarial robustness, compared to the \emph{Young} group. Looking at the cardinality distribution, we can see that the \emph{Young} is much larger than the \emph{Old} . Looking at the Gender attribute, the distributions of adversarial deviations among \emph{Women} are comparatively higher (higher median, comparable variance) compared to those among \emph{Men}, suggesting a relatively lower level of robustness within the \emph{Women} subgroup. However, \emph{Men} has a lower cardinality compared to \emph{Women}.

\emph{Conclusion:} From this experiment, we can infer that while imbalances in subgroup cardinalities do impact adversarial robustness, they do not singularly determine robustness levels. In one instance, the most resilient group (men) has relatively low cardinality, whereas in another instance, the most resilient group (youth) has comparatively higher cardinality. Other factors, such as the heterogeneity of the group, may contribute to a specific subgroup having low robustness despite its higher cardinality. We plan to further investigate this aspect in the future.
\subsubsection{Adversarial robustness across intersectional subgroups}
In Figure \ref{fig:instability_boxplot1} we show the adversarial deviations for the four subgroups based on gender and age for various $\beta$ values, alongside their respective cardinalities. 
In the adversarial deviations plot of the vanilla VAE ($\beta$=1), we observe that the groups \emph{young men} and \emph{young women} exhibit higher robustness compared to the groups \emph{old men} and \emph{old women}, respectively. In particular, the variances are higher for the \emph{old} subgroups. The \emph{old women} also have the higher median.
The \emph{young men} and \emph{young women} subgroups are the two best represented in the population (see bottom right). The group \emph{old women} depicts higher deviation values and greater variance, followed by \emph{old men}. Notably, these subgroups are the least represented in the population.

\textbf{The effect of $\beta$:} As  $\beta$ increases to 5.0 (top right), the distributions of adversarial deviations for all subgroups decrease correspondingly, indicating increased robustness. However, when the VAE was trained with an even higher value of $\beta = 10$, the adversarial robustness of all subgroups deteriorated compared to $\beta = 5$ and became as suboptimal as the vanilla VAE. The \emph{old women} group appears to be the most affected, showing significantly higher variability of adversarial deviations at $\beta = 10$. Despite the overall improvement in robustness for $\beta = 5$, the relatively higher adversarial deviation (comparatively low robustness) of the \emph{old women} group persists. Subgroup cardinalities indicate that the representation of the \emph{old women} subgroup is not sufficient for the model to ensure smooth and stable embeddings.

\emph{Conclusion:} Subgroup imbalances, which become more pronounced in the intersectional case, play a role in adversarial robustness. The disentanglement parameter $\beta$ affects the robustness even of imbalanced subgroups, but it requires an optimal value in our case, $\beta=5$. Despite the positive impact of $\beta$ on robustness, the relative robustness inequality persists even with optimal $\beta$. For example, the subgroup \emph{old women} remains relatively the least robust even after $\beta$ regularization.

\subsection{Qualitative analysis}
\subsubsection{Analysis of adversarial reconstruction losses}
\label{sec:expQualitative}
\begin{figure}
    \centering
    \begin{subfigure}[b]{1.0\textwidth}
        \includegraphics[width=0.90\textwidth]{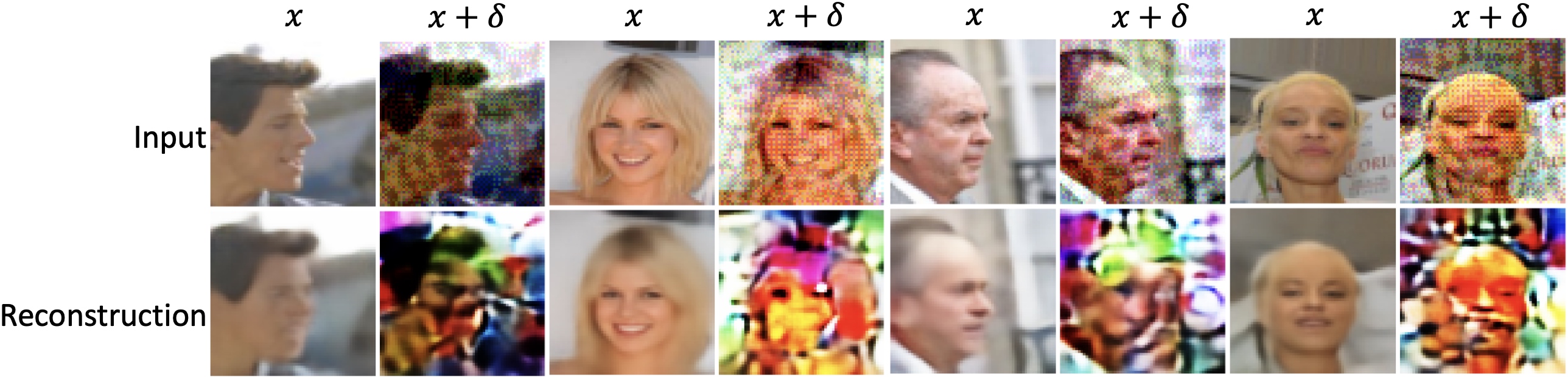}
        \caption{$\beta$=1}\label{fig:max_dam_qual_beta1}
    \end{subfigure}
    \begin{subfigure}[b]{\textwidth}
        \includegraphics[width=0.90\textwidth]{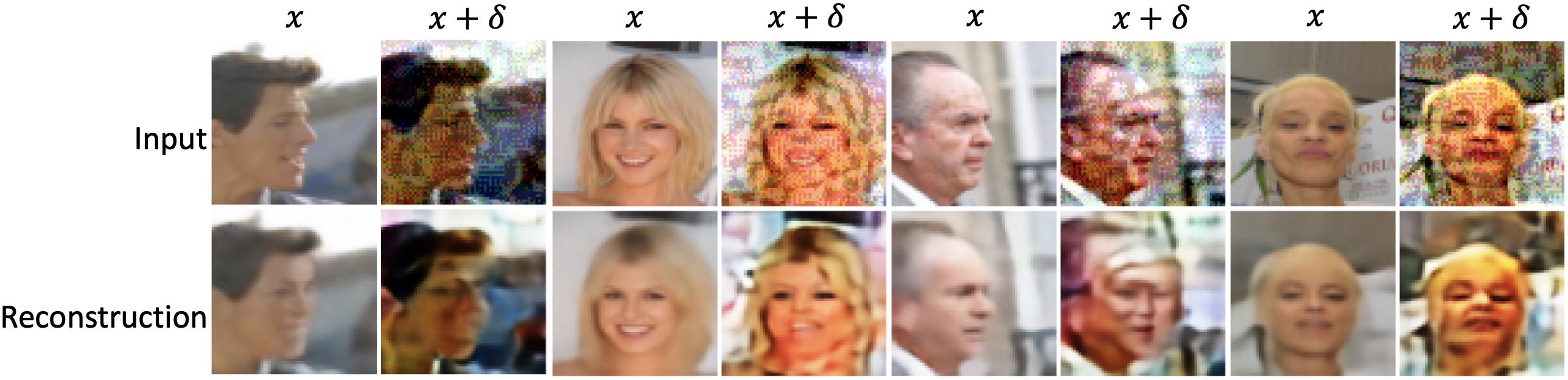}
        \caption{$\beta$=5}\label{fig:max_dam_qual_beta5}
    \end{subfigure}
    \begin{subfigure}[b]{\textwidth}
        \includegraphics[width=0.90\textwidth]{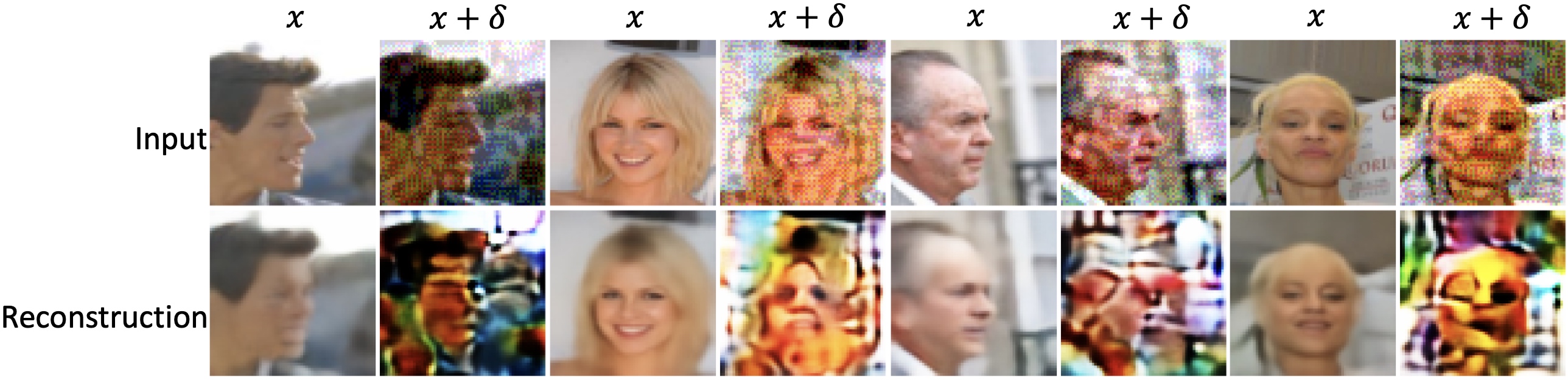}
        \caption{$\beta$=10}\label{fig:max_dam_qual_beta10}
    \end{subfigure}
    \caption{\small Inputs and reconstructions for normal ($x$) and perturbed samples ($x+\delta$) from the groups \emph{young men} (columns 1 \& 2), \emph{young women} (columns 3 \& 4), \emph{old men} (columns 5 \& 6),  and \emph{old women} (columns 7 \& 8) with highest adversarial deviation.}
    \label{fig:max_dam_qual}
\end{figure}

We select one sample from each subgroup (\emph{young men}, \emph{old men}, \emph{young women}, \emph{old women}): one causing maximum damage at the vanilla VAE output and generate maximum damage attack on the selected samples from all the subgroups with higher $\beta$-VAE modes. To compare reconstruction quality and adversarial deviations across $\beta$-VAEs, we plot these samples and observe their resistance to adversarial attacks with higher $\beta$ values ($\beta=5$ and $\beta=10$).
Figure \ref{fig:max_dam_qual} 
shows inputs and reconstructions from various $\beta$-VAE models for both normal and adversarial samples. These samples demonstrated maximum damage in each subgroup against the vanilla VAE ($\beta=1$) and illustrate the evolution of their resistance to attacks in other $\beta$-VAE modes.
\ref{fig:max_dam_qual_beta1} (for $\beta=1$), \ref{fig:max_dam_qual_beta5} (for $\beta=5$), and \ref{fig:max_dam_qual_beta10} (for $\beta=10$). 
The reconstruction quality of adversarial inputs is superior for $\beta$-VAEs with $\beta=5$ compared to those with $\beta=1$ and $\beta=10$. Figure \ref{fig:max_dam_qual} shows that adversarial reconstructions are more impaired for \emph{old women} and \emph{old men} than for \emph{young women} and \emph{young men}. Comparision between samples at $\beta=1$ and $\beta=10$ is difficult as they appear to be almost equally damages. Figure \ref{fig:max_dam_qual_beta5} demonstrates improvement in adversarial reconstructions with $\beta=5$ for all groups, especially preserving facial structure and attributes for \emph{young women} and \emph{young men} compared to \emph{old women} and \emph{old men}. Specifically, the \emph{young men} sample shows substantial improvement over the \emph{old men} sample.
Overall, certain subgroups exhibit higher adversarial deviations, indicating greater susceptibility to attacks and lower robustness.
During the experiments, we also observed a tendency for some minority group samples to be reconstructed as majority group samples even with optimal latent space regularization in the $\beta$-VAE ($\beta=5$). 
This poses a concern if $\beta$-VAE is used to mitigate adversarial attacks on a downstream classifier. 
Thus, we further investigate this tendency.
\subsubsection{Analysis of subgroup switching tendencies in adversarial reconstructions of minority samples}
\begin{figure}
    \centering
    \includegraphics[width=1.0\textwidth]{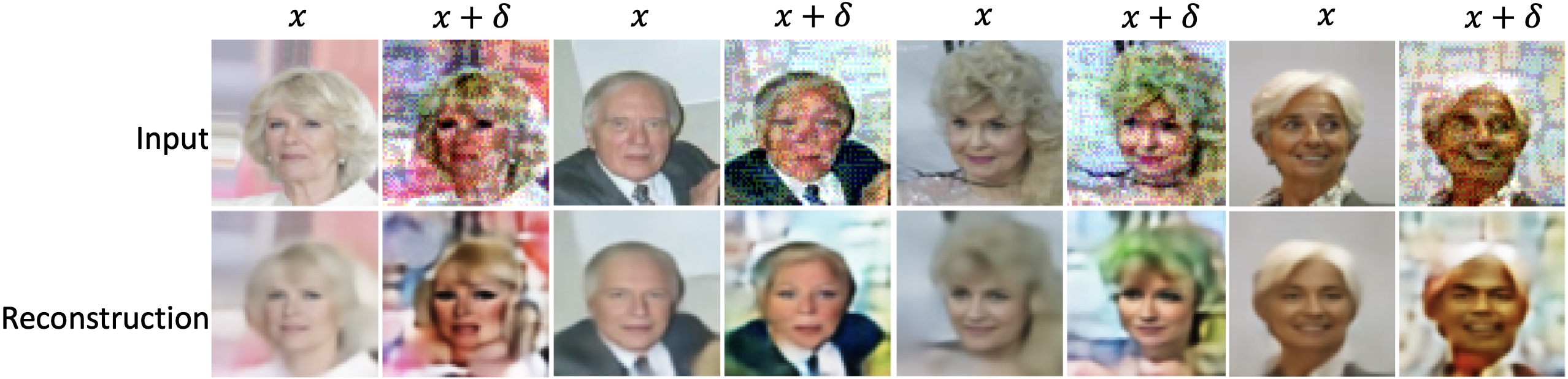}
    \caption{\small Inputs and reconstructions of samples that show tendency of subgroup switching in their reconstructions for $\beta$-VAE with $\beta$= 5.0}   
    \label{fig:group_switch}
\end{figure}

\begin{figure}
    \centering
    \begin{subfigure}[b]{0.32\textwidth}
        \includegraphics[width=\textwidth]{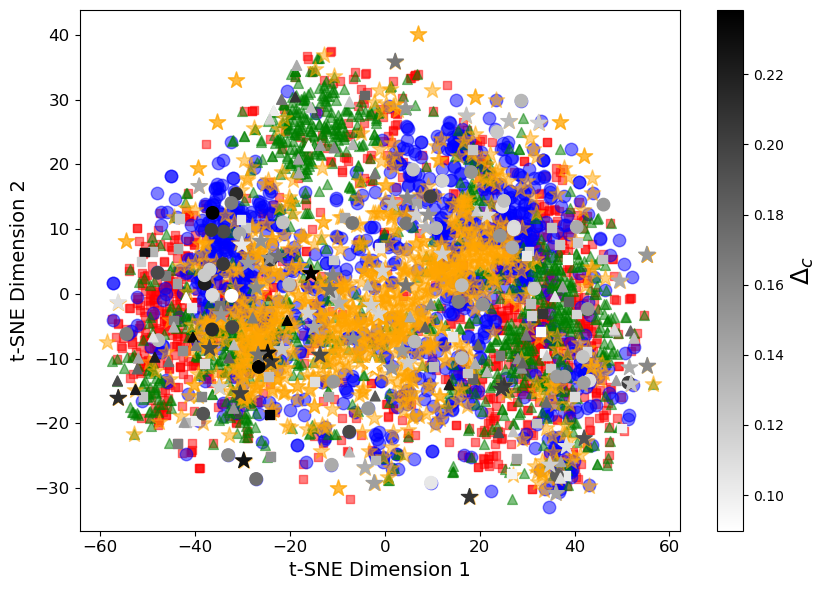}
    \end{subfigure}
    \begin{subfigure}[b]{0.32\textwidth}
        \includegraphics[width=\textwidth]{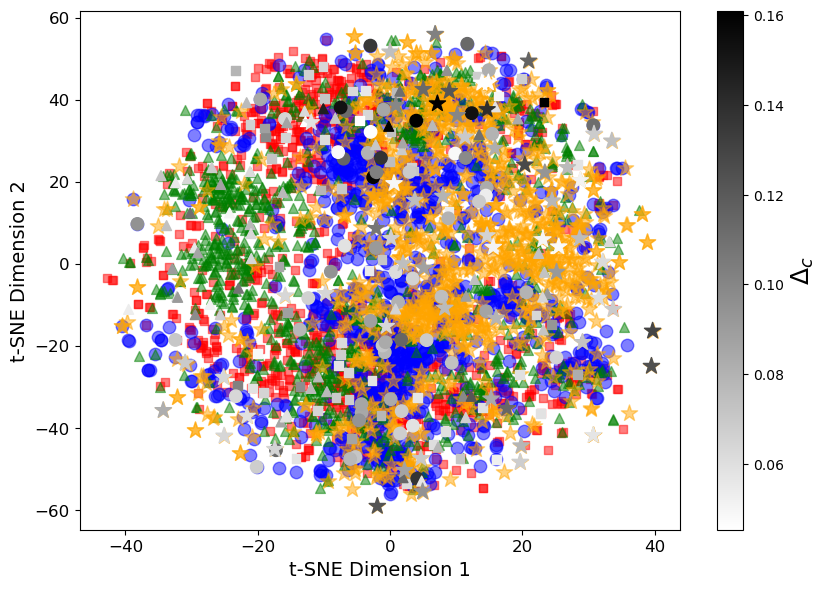}
    \end{subfigure}
    \begin{subfigure}[b]{0.32\textwidth}
        \includegraphics[width=\textwidth]{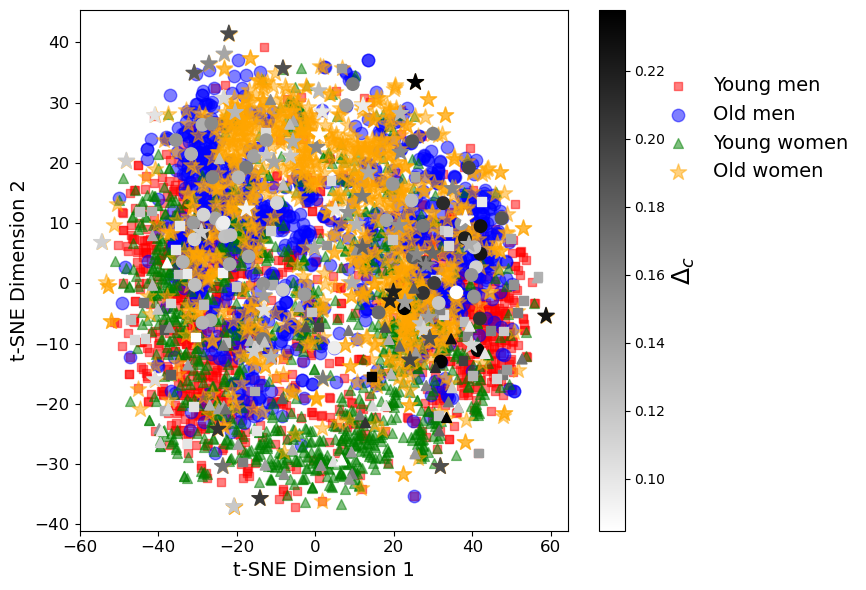}
    \end{subfigure}
    \caption{\small Embeddings of dataset subgroups along with  samples chosen for adversarial robustness evaluation for $\beta$-VAE with $\beta$=1.0 (top left), $\beta$-VAE with $\beta$=5.0 (top right) , $\beta$-VAE with $\beta$=10.0 (bottom). The gray-black scale color bar indicates the values of adverarial deviation. higher adversarial deviations $\Delta_c$ is observed for the subgroups \emph{old men} and \emph{old women}}   
    \label{fig:embedding_defects}
\end{figure}

Despite $\beta$-VAEs with an optimal $\beta=5$ minimize adversarial deviation, we observe that adversarial reconstructions often resemble those of majority groups in the dataset. Figure \ref{fig:group_switch} qualitatively shows $\beta$-VAE ($\beta=5$) adversarial reconstructions of \emph{old women} and \emph{old men} resembling the majority group.
To quantitatively investigate, we trained CNN classifiers for predicting gender and age (\emph{young}/\emph{old}) on the CelebA dataset, evaluating them on 60 samples from each subgroup. Tables \ref{tab:gender}  and \ref{tab:age} display the classifiers' prediction accuracy on direct images, reconstructions, and adversarial reconstructions for all three $\beta$-VAE modes.
Table \ref{tab:gender} shows the lowest gender prediction accuracy for \emph{old women} across all inputs. Notably, $\beta$-VAE ($\beta=5$) adversarial reconstructions inputs to gender classifier have higher prediction accuracy than those from vanilla-VAE ($\beta=1$). Table \ref{tab:age} indicates that the greatest reduction in age prediction accuracy for adversarial reconstructions, when used as inputs for the age classifier, is observed for the \emph{old women} group. This supports the observation that $\beta$-VAE ($\beta=5$) adversarial reconstructions of some of the minority group samples resemble the majority group.
We visualized latent representations of different subgroups using t-SNE on samples selected for adversarial robustness evaluation (Figure \ref{fig:embedding_defects}). Samples with maximum output damage were often in mixed or peripheral neighborhoods in latent space, while those with minimal damage were surrounded by similar subgroup points. Figure \ref{fig:embedding_defects} confirms that \emph{old women} and \emph{old men} samples suffered the most output damage from attacks.
\begin{figure}
    \includegraphics[width=\textwidth, height=5.5cm]{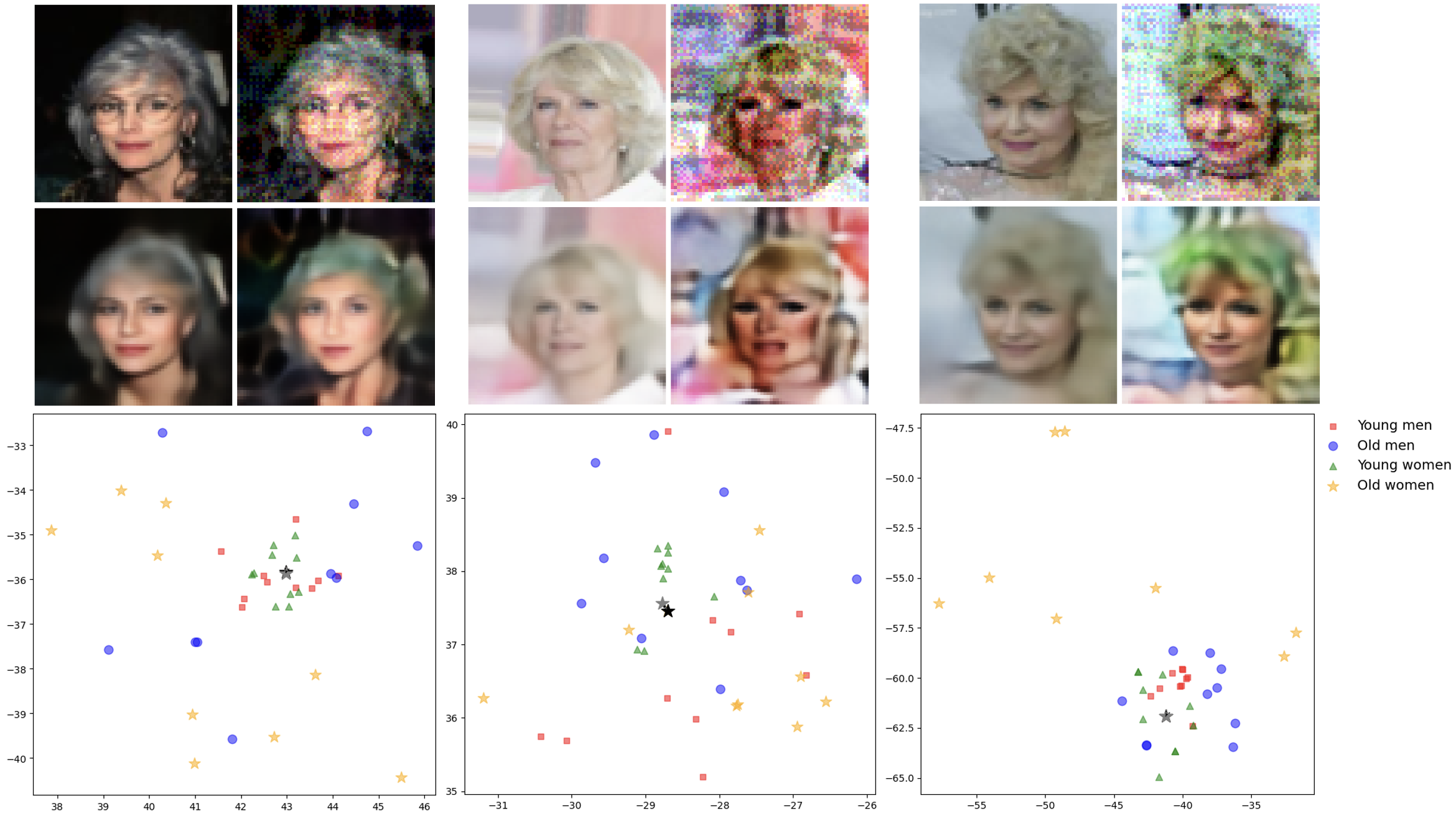}
    \caption{Pull effect on subgroup \emph{old women}: black star point indicates unperturbed embedding, gray dot indicates adversarial embedding}   
    \label{fig:old_women_pull}
\end{figure}
\begin{figure}
    \includegraphics[width=\textwidth, height=5.5cm]{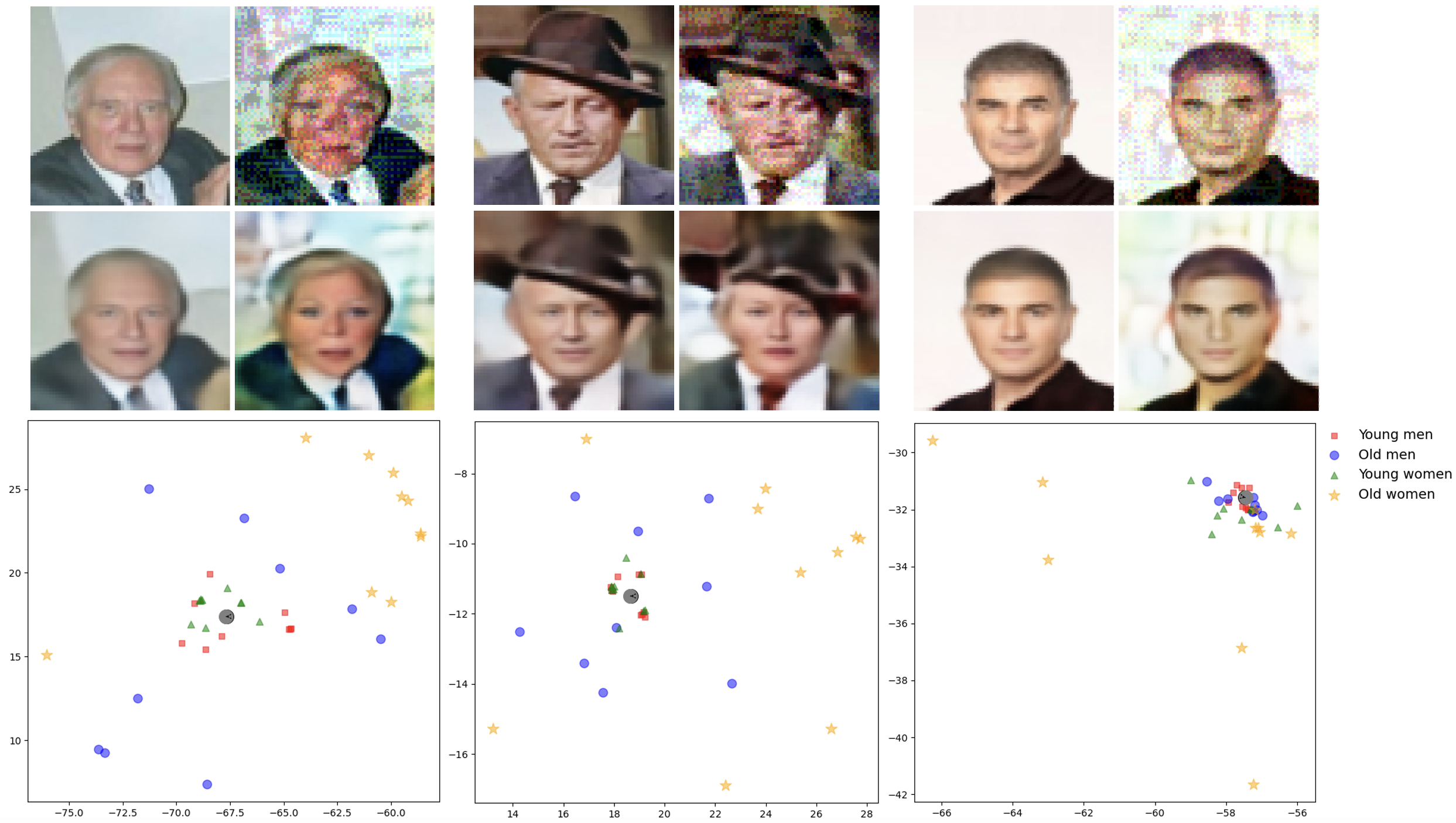}
    \caption{\small Pull effect on subgroup \emph{old men}: black dot indicates unperturbed embedding, gray dot indicates adversarial embedding}   
    \label{fig:old_men_pull}
\end{figure}
\begin{table}[h!]
\caption{The table shows the gender classifier prediction accuracy for 60 samples per subgroup across different input types and models. Accuracy is reported for direct input to the gender classifier(column 2), direct reconstructions from all $\beta$-VAEs modes (columns 3, 4, 5) and adversarial reconstructions from all $\beta$-VAE modes  (columns 6, 7, 8).}
\centering
\begin{tabular}{c@{\hskip 0.5cm}c@{\hskip 0.5cm}c@{\hskip 0.5cm}c@{\hskip 0.5cm}c@{\hskip 0.5cm}c@{\hskip 0.5cm}c@{\hskip 0.5cm}c}
\hline
subgroup & direct input & \multicolumn{3}{c@{\hskip 0.5cm}}{direct reconstruction} & \multicolumn{3}{c}{adversarial reconstruction} \\
\cmidrule(lr){3-5} \cmidrule(lr){6-8}
 &  & $\beta_{1}$ & $\beta_{5}$ & $\beta_{10}$ & $\beta_{1}$ & $\beta_{5}$ & $\beta_{10}$ \\
\hline
Young men  & 0.9500 & 0.95 & 0.9333 & 0.9333 & 0.6333 & 0.8167 & 0.8667 \\
Old men  & 0.9833 & 0.9833 & 0.9833 & 0.9667 & 0.7833 & 0.8167 & 0.8833\\
Young women  & 0.9667 & 0.9833 & 0.9667 & 0.9667 & 0.7333 & 0.9333 & 0.9161 \\
Old women & 0.9000 & 0.8667 & 0.90 & 0.85 & 0.5833 & 0.7833 & 0.7667 \\
\hline
\end{tabular}
\label{tab:gender}
\end{table}

\begin{table}[h!]
\caption{The table shows the age classifier prediction accuracy for 60 samples per subgroup across different input types and models. Accuracy is reported for direct input to the age classifier(column 2), direct reconstructions from all $\beta$-VAEs modes (columns 3, 4, 5) and adversarial reconstructions from all $\beta$-VAE modes  (columns 6, 7, 8).}
\centering
\begin{tabular}{c@{\hskip 0.5cm}c@{\hskip 0.5cm}c@{\hskip 0.5cm}c@{\hskip 0.5cm}c@{\hskip 0.5cm}c@{\hskip 0.5cm}c@{\hskip 0.5cm}c}
\hline
subgroup & direct input & \multicolumn{3}{c@{\hskip 0.5cm}}{direct reconstruction} & \multicolumn{3}{c}{adversarial reconstruction} \\
\cmidrule(lr){3-5} \cmidrule(lr){6-8}
 &  & $\beta_{1}$ & $\beta_{5}$ & $\beta_{10}$ & $\beta_{1}$ & $\beta_{5}$ & $\beta_{10}$ \\
\hline
Young men  &  0.7333 & 0.6667 & 0.6500 & 0.7000 & 0.4500 & 0.6833 & 0.6333 \\
Old men  & 1.000 & 0.9667 & 1.000 & 1.0000 & 0.8000 & 0.9667 & 0.9667\\
Young women  & 0.900 & 0.900 & 0.9167 & 0.9000 & 0.6167 & 0.8833 & 0.8833 \\
Old women & 0.9833 & 0.8833 & 0.9000 & 0.8833 & 0.7167 & 0.7667 & 0.7833 \\
\hline
\end{tabular}
\label{tab:age}
\end{table}

\subsection{Demostration of pull tendencies towards majority subgroups}
To further investigate, we visualize the embedding shift due to attacks and the neighborhood around the attacked sample embedding in Figures \ref{fig:old_men_pull} and \ref{fig:old_women_pull}. These figures illustrate samples from \emph{old women} and \emph{old men} subgroups that appeares to be switched to another subgroup in the reconstruction due to attacks against $\beta$-VAE with $\beta=5$. We selected a few samples from the \emph{old women} and \emph{old men} subgroups, plotted their unperturbed embedding and the shifted embedding due to attack, along with their 10 nearest neighbors from each subgroup choosing from the whole dataset embedding. The plots reveal that the embedding for \emph{old women} adversarial sample is located in a neighborhood of \emph{young women}. Similarly, the adversarial embedding of \emph{old men} samples are found in neighborhoods of \emph{young men} or \emph{young women}, influencing their reconstructions.

In Figures. \ref{fig:old_women_pull} and \ref{fig:old_men_pull}, it is clear that the direct reconstructions of the samples were not influenced by the majority samples in the neighborhood. However, a slight shift in the embedding due to adversarial perturbation causes the reconstruction to be heavily influenced by the surrounding majority group embeddings. This indicates that certain subgroups in the dataset, due to underrepresentation or specific attributes, can lead to non-smooth embeddings with defective latent manifold topology, potentially driven by the majority subgroup. These defective embeddings, when surrounded by majority subgroup samples, result in higher susceptibility to attacks, reduced robustness, and an increased tendency for reconstructions to be influenced by the majority subgroup samples in the neighborhood.

\section{Conclusion}\label{sec:conclusions}
Our study highlights the importance of robustness in representation learning across various (sub)groups to ensure ``fairness'' aiming for comparable adversarial robustness levels across all subgroups.
Despite the widespread adoption of autoencoders for representation learning in CV and their overall good performance, biases persist, especially against minority subgroups like \emph{old women}, reflecting issues of fairness and inclusivity. 
We found that while the representation of minority subgroups in training data significantly influences biases, simply increasing dataset size doesn't always address disparities. This suggests that a better notion of representativeness beyond cardinality is needed. Moreover, we discovered that enhancing disentanglement in the latent space of VAEs can improve fairness. This suggests the potential of deliberate efforts to promote disentanglement in VAE architectures by separating factors related to protected attributes from other variables, 
which we look to explore in future. However, disentanglement alone doesn't offer a panacea solution, as observed for the \emph{old women} subgroup in our experiments. 
Our research underscores the need for nuanced approaches to effectively mitigate biases, which may include enhancing representation for small subgroups as well implementing disentanglement.

\section{Acknowledgements}\label{sec:acknowledgements}
This research work received fund from the European Union under the Horizon Europe MAMMOth project, Grant Agreement ID: 101070285, and also supported by the EU Horizon Europe project STELAR, Grant Agreement ID: 101070122


%
%
%
\bibliographystyle{splncs04}
\bibliography{bias}
%




\end{document}

%% file: intro.tex
Autoencoders (AEs) have emerged as a versatile method in machine learning (ML) for various tasks such as dimensionality reduction \cite{fournier2019empirical}, representation learning \cite{tschannen2018recent} and data generation \cite{wan2017variational}. 
The widespread adoption of AEs, even in critical applications,  naturally raises concerns about their fairness in performance across different demographic subgroups and their robustness to adversarial attacks.

Despite significant progress in this area, it has become increasingly apparent that these models often learn biased representations with respect to protected attributes, such as gender or age, exhibiting discriminatory biases against minority subgroups like old women in various downstream tasks~\cite{wu2022fair-vae}. A contributing factor to this issue is the inherent biases present in the training set images~\cite{fabbrizzi2022Ntoutsisurvey}, which often feature a high disparity in representation towards minority subgroups. This disparity not only highlights issues of fairness and inclusivity but also raises critical ethical concerns about the deployment of such technologies in diverse societies.
Autoencoders are also not infallible; they are particularly susceptible to well-crafted input samples \cite{gondim2018adversarial}.
The adversarial samples, characterized by minimal, humanly imperceptible perturbations, can deceive otherwise high-performing deep learning models. 
This vulnerability poses significant challenges to the integrity and reliability of AEs on  applications in critical domains.
Variational Autoencoders (VAEs)~\cite{kingma2013vae,higgins2017beta} have emerged as a robust alternative to their deterministic counterparts (vanilla AEs), demonstrating a higher resilience to input perturbations, especially those stemming from adversarial attacks. 
Nonetheless, adversaries have devised methods to exploit the resilience of VAEs by introducing minor input perturbations~\cite{yuan2019adversarial} designed to elicit substantial changes in the encoding process.  
While the robustness of models against such adversarial strategies has been extensively studied~\cite{madry2017robust,carlini2019robust,olivier2023robust}, a gap remains in the comprehensive evaluation of model robustness across different demographics and whether there are differences in the performance.

In this study, we aim to provide a comprehensive evaluation of the robustness of VAEs against non-targeted adversarial attacks across various demographic subgroups. Our initial question (Q1) examines whether there are robustness disparities among these subgroups. The second part of our research (Q2) aims to understand the factors contributing to these disparities. We evaluate the adversarial robustness of different subgroups and examine the effect of the latent space disentanglement regularization parameter on subgroup robustness inequality. Although the $\beta$-VAE with an optimal regularization parameter reduced robustness disparities, it caused many samples to resemble majority class samples. We confirmed this by analyzing variations in prediction accuracy of downstream gender and age classifiers. To understand this behavior, we explored the latent space neighborhood of samples prone to subgroup switching reconstruction, identifying lack of embedding smoothness as a potential cause. 

%% file: related.tex
The evaluation of adversarial robustness involves generating adversarial attacks against the target model for specific samples and assessing the difficulty by quantifying the perturbation needed to induce incorrect predictions or reconstructions \cite{willetts2019improving}. Latent space attacks, as described by \cite{tabacof2016adversarial,kos2018adversarial}, optimize perturbations to minimize the KL divergence between latent distributions of perturbed and actual samples. \cite{barrett2022certifiably} proposed a modified latent space attack, constraining the perturbation norm by a constant instead of penalizing it within the adversarial objective. \cite{gondim2018adversarial} introduced a general targeted adversarial objective for attacking any autoencoder architecture. \cite{cemgil2019adversarially} suggested generating attacks by maximizing the Wasserstein distance between latent maps of input data and their adversarially perturbed counterparts, using the projected gradient descent method. \cite{camuto2021towards} presented the maximum damage attack for generating untargeted adversarial perturbations. While targeted attacks focus on producing a predefined output, untargeted attacks aim to make the autoencoder's output as different as possible from the original input, allowing the noise optimizer to explore a variety of perturbations that degrade performance. Thus, we choose untargeted adversarial attacks to evaluate robustness.

Given the widespread application of VAEs \cite{fournier2019empirical,petscharnig2017dimensionality} across various fields \cite{pratella2021survey} and being trained on datasets containing diverse human faces \cite{liu2015faceattributes}, it is essential to investigate VAE robustness in intersectional subgroups. These systems, may underperform for subgroups with specific facial attributes or those under-represented in the dataset, leading to biased performance. Despite extensive research, there has been limited investigation into comparing the adversarial robustness of different data subgroups. Therefore, this study aims to investigate the adversarial robustness of $\beta$ variants of Variational Autoencoders across intersectional subgroups.

%% file: background.tex
We assume an unsupervised learning scenario where a VAE model is used to learn a compact representation of high-dimensional image data. The problem formulation is presented in Section~\ref{sec:problem}, basic concepts on VAE models and adversarial attacks are presented in Section~\ref{sec:betaVAE} and Section~\ref{sec:adversarialAttacks}, respectively.

\subsection{Problem setup}
\label{sec:problem}
We assume an image dataset $\mathcal{I}=\{x_i,s_i\}$ , consisting of images $x_i\in \mathbf{R}^{N}$, and a vector of $k$ protected  values $s_i= [s_{i,1},\cdots, s_{i,k}]$, where each $s_{i,j}$ describes a demographic membership of $x_i$ (e.g. `Female') w.r.t. a protected attribute $s_{i,j}$ (e.g. `Sex').  For simplicity, we assume the protected attributes to be binary: $\forall  j=1,\cdots,k,  s_{i,j}\in \{g^j,\overline{g^j}\}$, where $g^j$ and $\overline{g^j}$ respectively represent the \emph{protected group} (e.g., female) and the \emph{non-protected group} (e.g., male). Further, the intersection of different protected attributes defines the so-called \emph{intersectional subgroups} or \emph{subgroups}, for short. For example, based on the binary protected attributes age=$\{$``young'', ``old''$\}$, and sex=$\{$``male'', ``female''$\}$, four different subgroups are formed: $\{$``young-female'',``young-male'', ``old-male'',``old-female''$\}$. The \emph{collection of subgroups} is denoted by  $\mathcal{SG}$ and defines as:
\begin{equation}\label{eq:sg}
  \mathcal{SG}=\{sg=s^1 \cap s^2\cap \cdots \cap s^k\mid~s^j\in \{g^j,\overline{g^j}\}, j=1,\cdots k\}\} 
\end{equation}  

Our goal is to learn a compact representation of the data using VAE models (c.f., Section~\ref{sec:betaVAE}). However, as the number of protected attributes increases, some subgroups may become smaller or even empty~\cite{roy2023facct} resulting in diverse qualities of representation learning across the subgroups. This variability could have direct consequences for adversarial robustness.
Our goal is to assess the adversarial robustness of VAE across various subgroups, and examine how the diverse qualities of representation
across the subgroups impact the vulnerability of each subgroup w.r.t. the adversarial attacks.

\subsection{Variational Autoencoders}
\label{sec:betaVAE}

Let $\mathcal{F}_{\phi, \theta}$ be a generative auto-encoder model utilizing a deep VAE~\cite{higgins2017beta} architecture, parameterised by an encoder with parameters $\phi$ that encodes any image $x\in \mathcal{I}$ into a compressed latent space representation $z\in \mathbf{R}^{M}$, $M<N$, and a decoder with parameters $\theta$ that decodes/reconstructs the image $x$ using the encoded representation $z$.
The aim for $\mathcal{F}_{\phi, \theta}$ is to learn $z$ as a generative latent representation  described by a conditional probability distribution $q_{\phi}(z|x)$, such that the predicted output space described by conditional probability $p_\theta(x|z)$ maximizes the likelihood of reconstructing the image $x$: $$\max\limits_\theta \mathbf{E}_{p(z)}[\log p_\theta (x|z)]$$ where $p(z)$ is the estimated prior of the latent representation $z$ drawn from standard normal distribution. 
The learning is accomplished using the following objective function:
\begin{equation}\label{eq.loss_vae}
    L(\theta, \phi) = E_{z \sim q_{\phi}(z|x)}[\log p_{\theta}(x|z)] - \beta \cdot  D_{KL}  (q_{\phi}(z|x) || p(z))
\end{equation}
where $\beta \geq 1 $ is a hyperparameter indicating the emphasis on the regularization term for latent space to be close to prior, with $\beta=1$ corresponding to the vanilla VAE~\cite{kingma2013vae}.  Increasing $\beta$ forces the model to learn more disentangled latent representation separating the distinct, independent and informative generative factors of variation in the data~\cite{wang2022disentangled,burgess2018understanding}. Studies have shown that learning a disentangled latent space can positively impact fairness of AEs~\cite{creager2019disentangle}, however it may also have a negative trade-off impact on the accuracy~\cite{kim2018disentangling} of the VAE. 

\subsection{Adversarial examples}\label{sec:adversarialAttacks}
Adversarial examples~\cite{yuan2019adversarial} in context of AEs for images are described as original input images with subtle modifications, typically imperceptible to humans. These slight alterations are carefully crafted to deceive image encoder, leading to error in the latent representation encoding, and henceforth, an incorrect reconstruction that differs highly from the original image. 

Formally, given an input image $x$, and an auto-encoder with learned reconstruction distribution $\mathbf{E}[\log p_\theta(x|z)]$, adversarial examples are defined as a subtle perturbation/modification $\delta$, which when added to the image as $x+\delta$, maximizes the gap between expected reconstruction:
\begin{equation}
\label{eq.loss_recons}
\begin{split}
      &\max\limits_\delta \|\mathbf{E}_{z' \sim q_{\phi}(z|x+\delta)}[\log p_\theta(x|z')]-\mathbf{E}_{z \sim q_{\phi}(z|x)}[\log p_\theta(x|z)]\|\\
      &\text{s.t.~~} \|\delta\|_p \leq c
\end{split}
\end{equation}
where $c$ is a hyperparameter bounding the norm of $\delta$ to ensure suitable perturbation intensity. This process falls under the category of \emph{non-targeted} and \emph{whitebox} adversarial attacks within the taxonomy of adversarial attacks~\cite{yuan2019adversarial}, where the adversary has access to the trained neural network model and tries to evade the system (e.g., Dodging~\cite{sharif2016dodge}) by learning an optimal noise using Equation~\ref{eq.loss_recons}. 

%% file: method.tex
Schematically, an overview of our approach for evaluating the adversarial robustness of a $\beta$-VAE  against non-targeted adversarial attacks across diverse subgroups is depicted in Figure~\ref{fig:ourOverview}.
It consists of two main components, attack generation and robustness evaluation, explained hereafter:
\begin{figure}[htbp]  
    \centering
\includegraphics[width=0.80\textwidth]{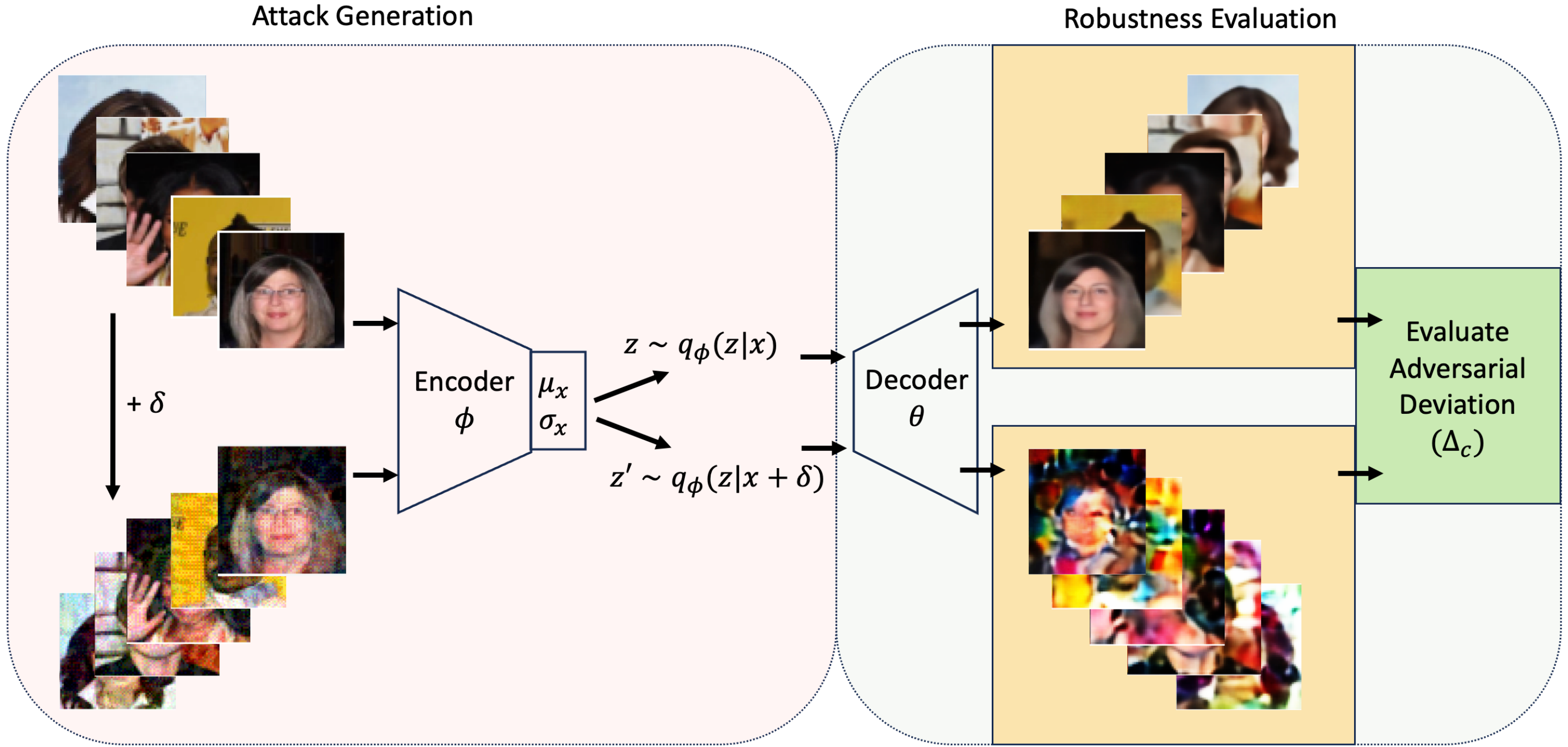}
    \caption{An overview of the  approach}\label{fig:ourOverview}
\end{figure}
\begin{enumerate}
    \item \textbf{Attack generation}: Attacks are generated as maximum damage attack instances to undermine the robustness of a $\beta$-VAE model (Section~\ref{sec:ourAdversarialAttacks}).
    \item \textbf{Robustness evaluation}: The robustness of a $\beta$-VAE model is evaluated across different (sub)groups against the generated attacks (Section~\ref{sec.ourEvaluation}).
\end{enumerate}

Given a trained $\beta$-VAE model $\mathcal{F}_{\phi,\theta}$, we intend to evaluate adversarial robustness of $\mathcal{F}_{\phi,\theta}$ across all the 
subgroups $sg \in \mathcal{SG}$. For each such subgroup $sg$ we sample a subset  
$\mathcal{I}^{(n)}_{sg} = \{ x_i| s_{i,j}=g_j,s_{i,k}=g_k, sg=g_j\cap g_k \}$ of $n$ randomly sampled points from the training dataset $\mathcal{I}$. 
We select the sample sets from the training data, to evaluate the model's vulnerability on the learned distribution itself, where it is expected to be most robust. Next, to generate the attacks, we learn an optimal perturbation as detailed in Section~\ref{sec:ourAdversarialAttacks} for each sample $x_i\in \mathcal{I}^{(n)}_{sg}$.
We use the learned perturbation to evaluate the model's robustness against the generated adversarial attack as described in Section~\ref{sec.ourEvaluation}, by measuring the deviation in the reconstruction of the adversarial input with that of the original input. Throughout the attack optimization, and the robustness evaluation stages, we consistently normalize the perturbed adversarial image before presenting it as input to the VAE. This practice mitigates the risk of generating trivial adversarial examples whose effects can be nullified by mere normalization.

\subsection{Maximum damage attack generation}
\label{sec:ourAdversarialAttacks}
In assessing VAE's robustness, the choice of the adversarial attack method plays a crucial role; we opt for maximum damage attacks~\cite{camuto2021towards}. This attack method aims to generate untargeted adversarial perturbations that inflict maximum damage on the reconstruction process of VAEs. 

The objective of a maximum damage attack is to maximize the discrepancy between the latent space representations of the original image $x$ and the perturbed image $x + \delta$, where $\delta$ represents a perturbation satisfying a given norm constraint. The main intuition is that by pushing the latent map away from the map of the unperturbed sample, the reconstruction loss at the VAE's output is expected to increase. 
Formally, the objective function is defined as follows:
\begin{equation}
    \label{max_damage}
    \underset{\substack{ \\ \delta : \|\delta\|_\infty \leq c}}{\text{arg max}} \, \| q_{\phi}(z | x + \delta) - q_{\phi}(z | x ) \|_2
\end{equation}
where $q_{\phi}(z | x)$ denotes the distribution of latent representations in VAE and $c$ is a hyperparameter specifying the bound on the norm of the perturbation $\delta$.

The utilization of the maximum damage attack across various protected groups and subgroups enables us to conduct rigorous assessments of VAE robustness, contributing to the understanding of the model's reliability and generalization capabilities 
under different demographic scenarios.




\subsection{Robustness evaluation for non-targeted adversarial attacks}
\label{sec.ourEvaluation}
Robustness to adversarial attacks is typically evaluated with adversarial accuracy loss~\cite{madry2017robust,carlini2019robust,olivier2023robust}. However, such evaluation set-up is typically applicable under a targeted adversary attack scenario. In VAE's for image learning, since there is no target label to learn, 
the main intent of the attacks is to deviate the model from proper reconstruction of the image. Thus, the deviation in the reconstructed output from the original reconstructed output~\cite{camuto2021towards} is used to test robustness.  


For each instance $x$  we evaluate the adversarial robustness of VAE $(\mathcal{F}_{{\phi},{\theta}})$ against attacks on $x$. 
We use the generated optimal distortion $\delta$ from the attack generation step (Section~\ref{sec:ourAdversarialAttacks}) to get the perturbed image $x+\delta$. Then, we provide both the original image $x$ and the perturbed image $x+\delta$ as inputs to VAE and we evaluate the deviation of the adversarial output from the output on the original image as: 
\begin{equation}
    \label{adv_div}
    \begin{aligned}
        &\Delta_c =  \|\mathbf{E}_{z' \sim q_{\phi}(z|x+\delta)}[\log p_\theta(x|z')]-\mathbf{E}_{z \sim q_{\phi}(z|x)}[\log p_\theta(x|z)]\|_2 
    \end{aligned}
\end{equation}
where $\Delta_c$ also referred as adversarial deviation is the measured $L_2$ norm distance between the reconstruction of the original image vs that of the perturbed image. Lower values of $\Delta_c$ correspond to higher robustness of the model against adversarial attacks on samples and vice versa.


%% file: dataset.tex
We experiment with the large-scale CelebFaces Attributes (CelebA) dataset \cite{liu2015faceattributes}, which consists of 202,599 celebrity images, each accompanied by 40 attribute annotations featuring diverse facial characteristics. 
The dataset is well-suited for this study due to its inclusion of various protected attributes and inherent imbalances in the population, which enables us to study robustness across various subgroups with varying cardinalities.

For this study, we consider Age and Gender, both binary, as the protected attributes to define the subgroups.  
An overview of the subgroups and their associated imbalances in cardinalities is shown in Figure~\ref{fig.subGcard}, with \emph{old women} comprising the smallest subgroup and \emph{young women} comprising the largest.

%% file: paper.bbl
\begin{thebibliography}{10}
\providecommand{\url}[1]{\texttt{#1}}
\providecommand{\urlprefix}{URL }
\providecommand{\doi}[1]{https://doi.org/#1}

\bibitem{barrett2022certifiably}
Barrett, B., Camuto, A., Willetts, M., Rainforth, T.: Certifiably robust variational autoencoders. In: AISTATS. PMLR (2022)

\bibitem{burgess2018understanding}
Burgess, C.P., Higgins, I., Pal, A., Matthey, L., Watters, N., Desjardins, G., Lerchner, A.: Understanding disentangling in $beta$-vae. arXiv preprint  (2018)

\bibitem{camuto2021towards}
Camuto, A., Willetts, M., Roberts, S., Holmes, C., Rainforth, T.: Towards a theoretical understanding of the robustness of variational autoencoders. In: AISTATS. PMLR (2021)

\bibitem{carlini2019robust}
Carlini, N., Athalye, A., Papernot, N., Brendel, W., Rauber, J., Tsipras, D., Goodfellow, I., Madry, A., Kurakin, A.: On evaluating adversarial robustness. arXiv preprint  (2019)

\bibitem{cemgil2019adversarially}
Cemgil, T., Ghaisas, S., Dvijotham, K.D., Kohli, P.: Adversarially robust representations with smooth encoders. In: ICLR (2019)

\bibitem{creager2019disentangle}
Creager, E., Madras, D., Jacobsen, J.H., Weis, M., Swersky, K., Pitassi, T., Zemel, R.: Flexibly fair representation learning by disentanglement. In: ICML (2019)

\bibitem{ehrhardt2022autoencoders}
Ehrhardt, J., Wilms, M.: Autoencoders and variational autoencoders in medical image analysis. In: Biomedical Image Synthesis and Simulation (2022)

\bibitem{fabbrizzi2022Ntoutsisurvey}
Fabbrizzi, S., Papadopoulos, S., Ntoutsi, E., Kompatsiaris, I.: A survey on bias in visual datasets. CVIU  (2022)

\bibitem{fournier2019empirical}
Fournier, Q., Aloise, D.: Empirical comparison between autoencoders and traditional dimensionality reduction methods. In: 2019 AIKE. IEEE

\bibitem{gondim2018adversarial}
Gondim-Ribeiro, G., Tabacof, P., Valle, E.: Adversarial attacks on variational autoencoders. arXiv preprint  (2018)

\bibitem{higgins2017beta}
Higgins, I., Matthey, L., Pal, A., Burgess, C.P., Glorot, X., Botvinick, M.M., Mohamed, S., Lerchner, A.: beta-vae: Learning basic visual concepts with a constrained variational framework. ICLR (Poster)  (2017)

\bibitem{karovaliya2015atm}
Karovaliya, M., Karedia, S., Oza, S., Kalbande, D.: Enhanced security for atm machine with otp and facial recognition features. Procedia Computer Science  (2015)

\bibitem{kim2018disentangling}
Kim, H., Mnih, A.: Disentangling by factorising. In: ICML (2018)

\bibitem{kingma2013vae}
Kingma, D.P., Welling, M.: Auto-encoding variational bayes. arXiv preprint arXiv:1312.6114  (2013)

\bibitem{kos2018adversarial}
Kos, J., Fischer, I., Song, D.: Adversarial examples for generative models. In: 2018 ieee security and privacy workshops (spw). IEEE

\bibitem{liu2015faceattributes}
Liu, Z., Luo, P., Wang, X., Tang, X.: Deep learning face attributes in the wild. In: Proceedings ICCV (2015)

\bibitem{madry2017robust}
Madry, A., Makelov, A., Schmidt, L., Tsipras, D., Vladu, A.: Towards deep learning models resistant to adversarial attacks. arXiv preprint arXiv:1706.06083  (2017)

\bibitem{olivier2023robust}
Olivier, R., Raj, B.: How many perturbations break this model? evaluating robustness beyond adversarial accuracy. In: ICML. PMLR (2023)

\bibitem{petscharnig2017dimensionality}
Petscharnig, S., Lux, M., Chatzichristofis, S.: Dimensionality reduction for image features using deep learning and autoencoders. In: Proceedings of the 15th international workshop on content-based multimedia indexing (2017)

\bibitem{pratella2021survey}
Pratella, D., Ait-El-Mkadem~Saadi, S., Bannwarth, S., Paquis-Fluckinger, V., Bottini, S.: A survey of autoencoder algorithms to pave the diagnosis of rare diseases. Int. J. Mol. Sci  (2021)

\bibitem{roy2023facct}
Roy, A., Horstmann, J., Ntoutsi, E.: Multi-dimensional discrimination in law and machine learning-a comparative overview. In: ACM FAccT (2023)

\bibitem{sharif2016dodge}
Sharif, M., Bhagavatula, S., Bauer, L., Reiter, M.K.: Accessorize to a crime: Real and stealthy attacks on state-of-the-art face recognition. In: Proceedings of the 2016 acm sigsac conference on computer and communications security (2016)

\bibitem{tabacof2016adversarial}
Tabacof, P., Tavares, J., Valle, E.: Adversarial images for variational autoencoders. arXiv preprint arXiv:1612.00155  (2016)

\bibitem{tschannen2018recent}
Tschannen, M., Bachem, O., Lucic, M.: Recent advances in autoencoder-based representation learning. arXiv preprint  (2018)

\bibitem{wan2017variational}
Wan, Z., Zhang, Y., He, H.: Variational autoencoder based synthetic data generation for imbalanced learning. In: 2017 SSCI. IEEE (2017)

\bibitem{wang2022disentangled}
Wang, X., Chen, H., Tang, S., Wu, Z., Zhu, W.: Disentangled representation learning. arXiv preprint  (2022)

\bibitem{willetts2019improving}
Willetts, M., Camuto, A., Rainforth, T., Roberts, S., Holmes, C.: Improving vaes' robustness to adversarial attack. arXiv preprint arXiv:1906.00230  (2019)

\bibitem{wu2022fair-vae}
Wu, C., Wu, F., Qi, T., Huang, Y.: Semi-fairvae: Semi-supervised fair representation learning with adversarial variational autoencoder. arXiv preprint  (2022)

\bibitem{yuan2019adversarial}
Yuan, X., He, P., Zhu, Q., Li, X.: Adversarial examples: Attacks and defenses for deep learning. IEEE transactions on neural networks and learning systems  (2019)

\end{thebibliography}
